\documentclass[letterpaper, 10 pt, conference]{ieeeconf}  

\IEEEoverridecommandlockouts                              

\usepackage{graphicx}
\usepackage[dvipsnames]{xcolor}

\title{\LARGE \bf
Autonomous Sensor Exchange and Calibration \\for Cornstalk Nitrate Monitoring Robot
}

\author{Janice Seungyeon Lee, Thomas Detlefsen, Shara Lawande, Saudamini Ghatge, Shrudhi Ramesh Shanthi, \\Sruthi Mukkamala, George Kantor, and Oliver Kroemer 
\thanks{$^{1}$J. S. Lee, T. Detlefsen, S. Lawande, S. Ghatge, S. R. Shanthi, S. Mukkamala, , G. Kantor and O. Kroemer are with Robotics Institute, School of Computer Science,
        Carnegie Mellon University, Pittsburgh, PA 15213, USA
        {\tt\small $\{$janicel2, tdetlefs, slawande, sghatge, srameshs, sruthim,  kantor, okroemer$\}$@andrew.cmu.edu}}%
}

\begin{document}

\maketitle
\thispagestyle{empty}
\pagestyle{empty}

\begin{abstract}

Interactive sensors are an important component of robotic systems but often require manual replacement due to wear and tear. Automating this process can enhance system autonomy and facilitate long-term deployment. We developed an autonomous sensor exchange and calibration system for an agriculture crop monitoring robot that inserts a nitrate sensor into cornstalks. A novel gripper and replacement mechanism, featuring a reliable funneling design, were developed to enable efficient and reliable sensor exchanges. To maintain consistent nitrate sensor measurement, an on-board sensor calibration station was integrated to provide in-field sensor cleaning and calibration. The system was deployed at the Ames Curtis Farm in June 2024, where it successfully inserted nitrate sensors with high accuracy into 30 cornstalks with a 77$\%$ success rate.

\end{abstract}

\section{INTRODUCTION}

Interactive sensors, such as tactile sensors and physical probes, are an important component for robots collecting data from their environment. The ability to acquire data during interactions allows a robot to observe information about objects that it would not be able to detect using only passive sensor observations. However, interactive sensors often experience extensive wear and tear that requires them to be replaced on a regular basis. This exchanging of sensors is often performed manually by a human operator, which limits the overall autonomy of the robot system and its ability to be deployed for extended periods of time. Exchanging sensors does not only involve replacing the sensor itself, but each sensor must also be calibrated and cleaned in the field for consistent measurements and extended use. Cleaning is particularly important for interactive sensors such as physical probes. 

In this paper, we explore the challenges of creating an autonomous sensor exchange and maintenance system for an agriculture robot monitoring cornstalks. The robot is tasked with forcefully pushing a sensor into cornstalks in the field to measure their nitrate readings. We use an existing sensor technology developed by Ali et al \cite{sensor}. The sensor has two electrodes printed on a PCB with electrochemical layers applied on top that converts nitrate concentration to voltage (see \cite{sensor} for details). Our focus is in creating the infrastructure needed to repeatedly deploy multiple of such sensors in the field without additional human intervention. 

Our approach includes three core components. First, a gripper was designed for engaging and disengaging with the sensor both mechanically and electronically. Secondly, a station was developed for removing expired sensors and housing replacement sensors in a precise and accessible manner using a physical funneling approach. Lastly, a calibration and cleaning station was developed for preparing new sensors and cleaning the sensor between samplings without creating cross contamination. Although our focus is on the nitrate sensor, many of the underlying principles of our system can also be adopted for other physical sensors.

\begin{figure}[t]
    \centering
    \includegraphics[width=0.95\linewidth]{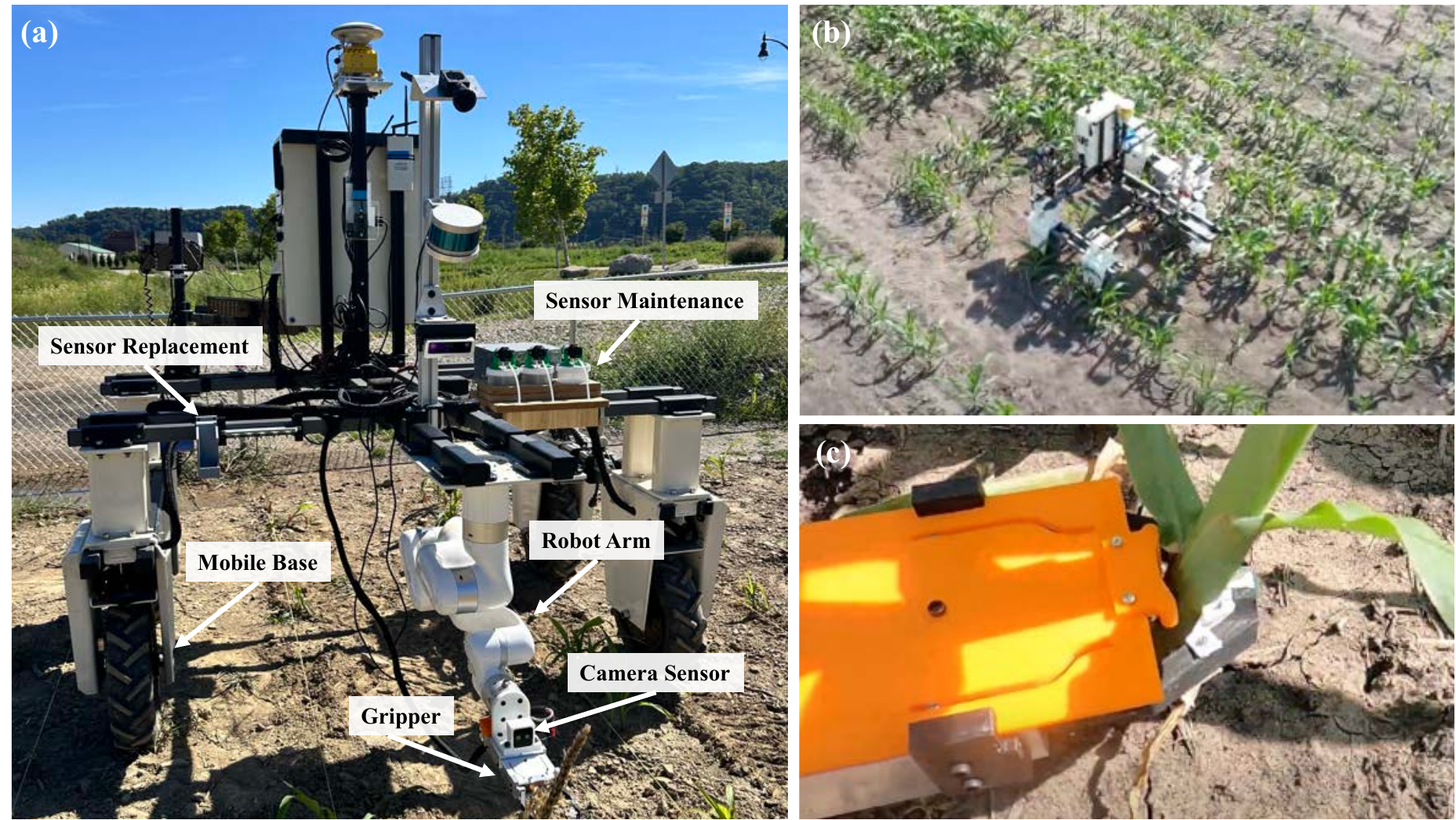}
    \caption{(a) Robot platform hardware overview. (b) Robot deployed in the cornfield (c) Gripper inserting sensor into stalk}
    \label{fig:robot_system}
    \vspace{-15pt}
\end{figure}

\begin{figure*}[t]
    \vspace{1.5mm}
    \centering
    \includegraphics[width=0.95\linewidth]{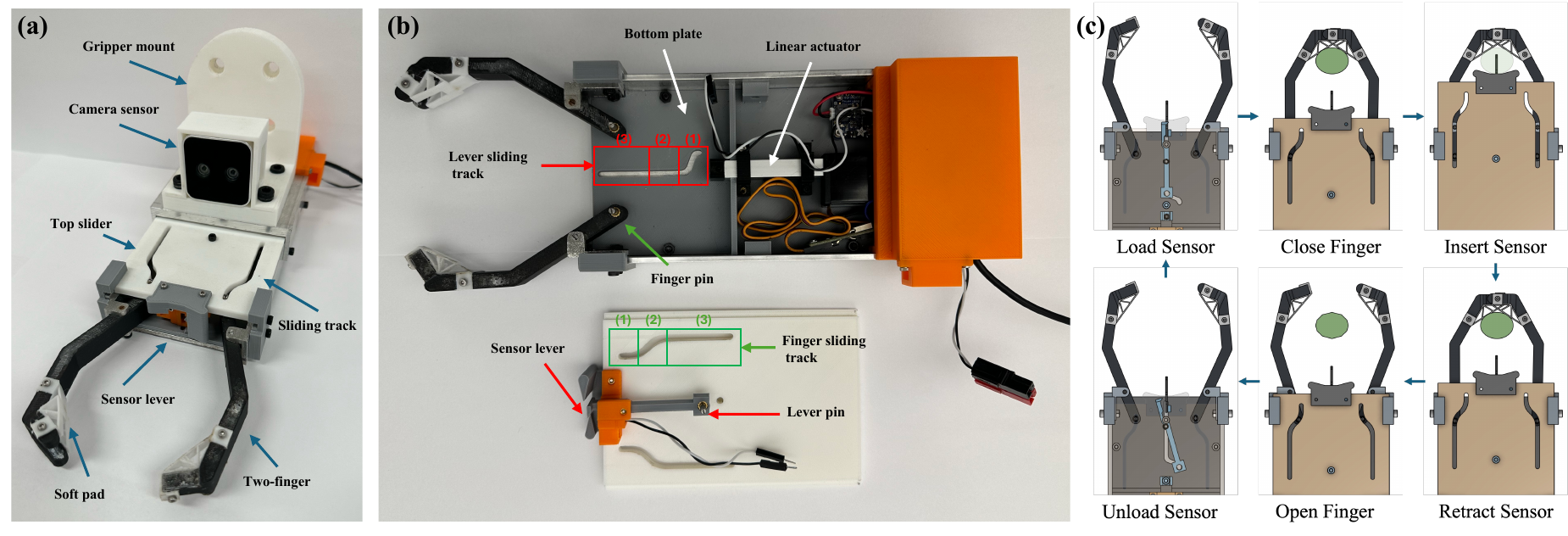}
    \caption{(a) Gripper uses two-finger grasp mechanism and has the camera facing forward. (b) Each proportion of the slide tracks for the sensor lever pin and the finger pins are for 1) moving the sensor lever 2) opening and closing finger 3) inserting and retracting the sensor. (c) The diagram shows the finger and sensor lever movement based on the extended length of the linear actuator.}
    \label{fig:gripper}
    \vspace{-15pt}
\end{figure*}

\section{RELATED WORK} \label{relatedwork}

\subsection{Sensors and Robots in Agriculture}
Precision agriculture has recently been using advancement of sensing technology and robotics to optimize resources and improve yields and profits \cite{pa, pa2}. A variety of sensing technologies are employed in agricultural applications, including vision-based sensing \cite{vision} for applications such as structural characterization \cite{sc1, sc2} and plant detection \cite{detect1, detect2}, as well as interactive sensors for soil \cite{soil1, sensor2} and crop monitoring \cite{sensor}. 
These sensors are integrated with robotic systems to improve autonomy for labor-intensive and repetitive agricultural tasks such as monitoring \cite{monitoring}, harvesting \cite{harvesting} and management \cite{management}. Our previous work \cite{pw1} deployed individual sensors and left them in the cornstalks. The next sensor then had to be manually reloaded. By contrast to previous works, our approach focuses on allowing the robot to autonomously replace, clean, and calibrate interactive sensors in the field. 

Grippers are crucial end-of-arm tools for robotic manipulation, facilitating interaction with environments and objects \cite{ag_grip}. Specially for grasping objects, two-finger designs are often used for their simplicity \cite{mason, tf}. The two-finger design can be actuated for inserting task utilizing a linear actuator and sliding tracks to enable both insert and grasp motion \cite{gripper1}. Gripper are often customized for tool-switching function, involving mechanical \cite{ts1} and electrical engagement \cite{ts2}. This research presents a novel gripper capable of reliable autonomous sensor switching while maintaining effective grasping and insertion capabilities.

Sensor exchange necessitates precise alignment and insertion, particularly for small-sized sensors. Guo et al. developed an intelligent robotic hand equipped with multiple small sensors for tolerant electronic connector mating \cite{prec0}. Gregorio et al. used computer vision and tactile sensor for wire terminal insertion \cite{prec1}.  Morgan et al. introduced combined vision-based object tracking, compliant manipulation, and learned hand models to accomplish tight-tolerance peg-in-hole insertions \cite{prec2}. For a mechanical solution, Nie et al. proposed a gripper design with an L-shaped finger that acts as a guide for aligning \cite{Lgripper}. Our research explores a novel mechanical solution using a simple and cost-effective funneling mechanism. 

The main contributions of this paper are:
\begin{itemize}
    \item Development of a custom gripper using a single-actuator coupled sliding mechanism for stalk grasping, sensor inserting, and sensor replacing
    \item Reliable high-precision sensor replacement for low-precision robot manipulator using cost-effective funneling mechanism
    \item On-board sensor calibration and cleaning mechanism for autonomous infield sensor maintenance. 
    \item Evaluation of the robot system on real-world cornstalks with multiple sensors.
\end{itemize}

\section{ROBOT SYSTEM OVERVIEW}ary application goal of this research is the development of a robotic system that is capable of inserting nitrate  asensor into multiple cornstalks, and]]]]\\while replacing and maintaining the sensors between insertions, for autonomous crop monitoring. The robotic system we developed consists of four subsystems - (i) a gripper for inserting sensors into cornstalks, (ii) a replacement mechanism for exchanging sensors, (iii) a sensor maintenance mechanism for calibrating and cleaning the sensors, and (iv) a visual-detection algorithm for detecting cornstalks and their width.

These subsystems are placed on a commercially available, customizable four-wheel AMIGA mobile base (see Fig. \ref{fig:robot_system}), which was adapted to fit the dimensions of the cornfield \cite{hefty}. At the testing site, the cornstalks were planted in rows with row spacing of 75 cm and were at the growth stage between V6 - V7 which has the average height of 60 cm. The dimension of the AMIGA base was set to be 150 cm x 86 cm (W x H) to provide clearance. With this configuration, the base can straddle two rows of cornstalks while the 6 D.O.F. xArm robot is positioned in the center of the rows. The robot operates with an Intel i9-processor with RTX4070 GPU and utilizes ROS framework.

\section{GRIPPER DESIGN} \label{gripper}

A gripper was developed to achieve sensor insertion and autonomous exchange utilizing two-finger design with soft pads and coupled sliding mechanism.

\subsection{Sensor Insertion Requirements} \label{sensor_insert_require}

To ensure accurate sensor readings, the sensor must be inserted at least $8.5$mm into the cornstalk's pith region. The nitrate sensor consists of a flat spike that is $5$mm wide, $12$mm long, and $1.6$mm thick. Two electrodes are located on one side of the spike along the central axis, $3$mm from the tip and $5.5$mm apart \cite{sensor}. The sensor needs to be inserted fully covering both of the electric nodes, which corresponds to an insertion depth of $8.5$mm. Also, the sensor must be inserted within the pith region of the cornstalk, which can be achieved by inserting the sensor $1.3$ cm to $2.5$ cm above the ground. The nitrate concentrate decreases by 4$\%$ for every centimeter deviation above the ground \cite{pith1}. Fulfilling these depth and height requirements will allow the robot to get accurate and consistent measurements.

The gripper serves multiple functions, including grasping cornstalks, inserting the sensor into the grasped stalk, and exchanging sensors. The gripper must be applicable to a range of stalk diameters (between 15mm and 35 mm \cite{corn}), grasping and centering the stalks for precise insertions. To work close to the ground, maximize the workspace in the crowded field, and not unduly burden the arm, the gripper should be compact and light. Additionally, the gripper must be designed to facilitate autonomous sensor exchange. We will go over the gripper design decisions used to address these requirements. 

\subsection{Gripper Mechanisms}

The gripper has an overall dimension of 9cm x 10cm x 28cm (HxWxD). As shown in Fig. \ref{fig:gripper}(a), the gripper has a two-finger design \cite{gripper1} for grasping, and each of the finger tips has a soft pad to help align the stalk to the center of the gripper. The camera is placed on top of the gripper facing forward for detecting cornstalks. The gripper uses a 50 mm stroke linear actuator from Actuonix to grasp stalks, and insert and exchange sensors. The main frames and the finger brackets that have large forces applied were fabricated using aluminum. Other parts, such as the fingers, lever and slider, that require specific customized features were 3D printed using PLA filament.

The two-finger gripper is designed to efficiently navigate to the target stalk in a straight line with the camera sensor facing the same direction. This approach reduces positional errors resulting from changing orientation or direction and the risk of colliding with adjacent stalks. On each of the finger-tips, a custom soft pad was attached to form a v-shape when grasping, which assists in aligning the stalk while being capable of grasping a wide range of stalk diameters. The soft pads is able to grasp relatively thick stalks since it is deformable, and for the thin stalks, the pads push the stalk towards the center during the grasping process. These soft pads were fabricated with TPU filament using a 3D printer. 

\begin{figure}[b]
    \centering
    \includegraphics[width=0.95\linewidth]{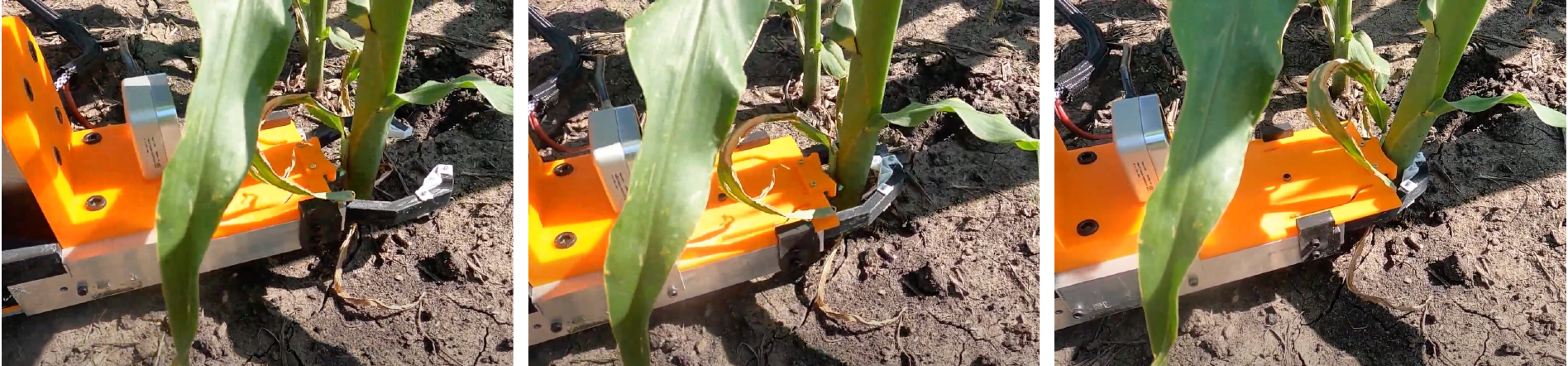}
    \caption{Gripper motion for grasping the stalk and inserting sensor.}
    \label{fig:gripper_motion}
\end{figure}

The gripper is actuated with one linear actuator that utilizes a coupled sliding mechanism. This mechanism enables three motions: (i) opening and closing the fingers for grasping stalks, (ii) inserting the sensor into the grasped stalk, and (iii) loading and unloading the sensor. As shown in Fig. \ref{fig:gripper}(b), there are two different sliding tracks where one on the bottom plate is for the sensor lever and the one on the top slider is for the finger movement. While the bottom plate remains stationary attached to the main frame, the top slider is connected to the linear actuator and slides forward. The finger pins attached to the inner end of the fingers slides through the finger tracks leading to the finger grasping and sensor insertion motions, as shown in Fig. \ref{fig:gripper_motion}. For the sensor lever motion, the pin at the end of the sensor lever moves along the lever pin track located on the bottom plate. As the lever pivots when the slider is moving forward, it hooks the sensor in the sensor slot located in the front of the top slider and unloads the sensor when the linear actuator retracts fully (see Fig. \ref{fig:sensor1}(c)). 

\section{SENSOR REPLACEMENT} \label{replace_mech}

One sensor is used for multiple insertions to collect readings across multiple cornstalks within a field. However, after multiple forceful penetrations into stalks, the chemical layers on top of the sensor is subject to potential damage. Thus, the sensor needs to be replaced with a new sensor periodically. We replace the sensor after every 5 plant insertions.

\subsection{Sensor Modification}

\begin{figure}[t]
    \vspace{1.5mm}
    \centering
    \includegraphics[width=0.95\linewidth]{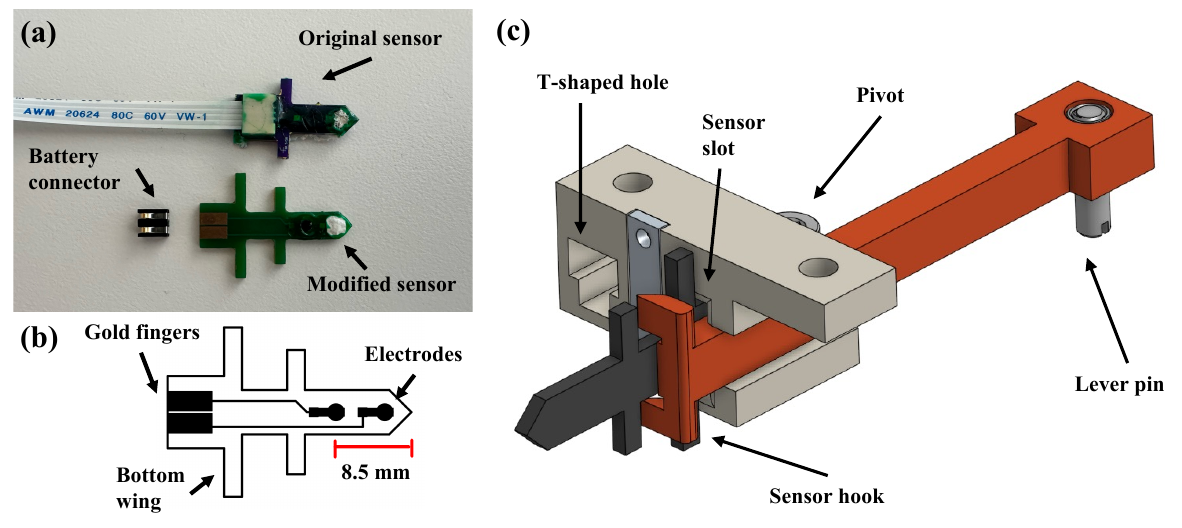}
    \caption{(a) Modified sensor compared with original sensor (b) Electrical diagram of the sensor (c) Sensor lever mechanism}
    \label{fig:sensor1}
    \vspace{-15pt}
\end{figure}

The sensor was modified to have contact-based electrical connections and an additional wing shape to allow it to be mechanically hooked and secured. The original nitrate sensor utilized a ribbon cable for electrical connection. However, to allow reliable loading and unloading by the gripper, the sensor was modified to incorporate a contact-based electrical interface using gold finger connections, as shown in Fig. \ref{fig:sensor1}(a). As the sensor is inserted in the sensor slot of the gripper, it makes contact-based connection with the battery connector, which interfaces with the on-board electronics for reading the sensor. Our modified sensor also has two wings, and the additional wing is used for the sensor lever to hook on to for a strong mechanical engagement when retracting the sensor from the sensor holder and the stalk. 

\subsection{Lever Mechanism for Loading Sensor}

As shown in Fig. \ref{fig:sensor1}(c), the gripper utilizes a lever mechanism to hook on to the sensor for a secure mechanical engagement. Inserting the sensor into corn and retracting are forceful tasks requiring 30N force, and thus, when the sensor is loaded, it needs to be held securely in place. The sensor slot is attached to the front end of the top slider and moves linearly along with the linear actuator. The lever pivots as the lever pin slides forward along the sensor lever sliding track in the bottom plate of the gripper (Fig. \ref{fig:gripper}(b)), securely hooking on the loaded sensor. When the linear actuator is retracted all the way, the lever opens up and the sensor slot moves inwards toward the gripper main frame. The bottom wing of the sensor then collides with the metal frame, which forces the sensor out of the slot. 

\subsection{Funnel Mechanism for Aligning Sensor}

\begin{figure}[t]
    \vspace{2mm}
    \centering
    \includegraphics[width=0.95\linewidth]{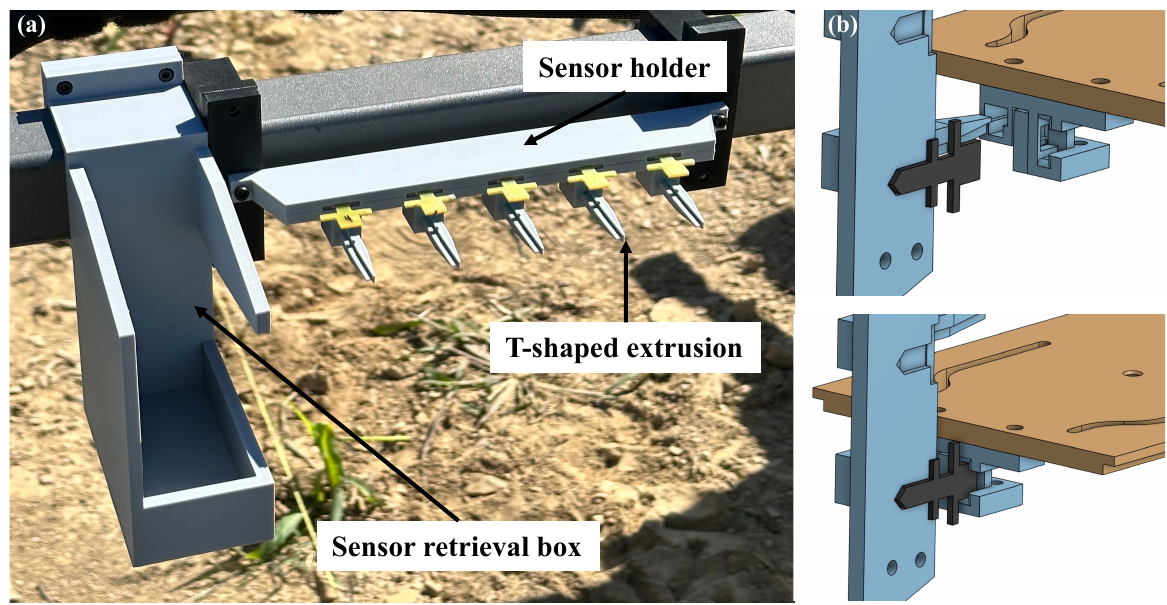}
    \caption{(a) The sensor replacement mechanism consists of the sensor holder with T-shaped extrusion and the sensor retrieval box. (b) The T-shaped extrusion funnels in to the T-shaped slot adjacent to the sensor slot.}
    \vspace{-15pt}
    \label{fig:replacement}
\end{figure}

When loading a new sensor, the sensor needs to be aligned accurately to the sensor slot on the gripper due to a submillimeter tolerance. The sensor slots have tight fit for the sensor because it establishes contact-based electrical connection and the lever mechanical engagement. The surface area of the electrical contacts on the sensor is 2.2mm x 5mm (WxH) which 
necessitates submillimeter precision for alignment. However, the sensor slot on the gripper is located on the top slider which has low-precision movements. To address this, a funneling mechanism was implemented for aligning the sensor closely to the sensor slot. 

The sensor replacement mechanism, placed on the right front bar of the AMIGA base (Fig. \ref{fig:replacement}(a)), consists of a sensor holder that holds 5 five sensor in a known location and a sensor retrieval box. The sensor holder has sensors placed in each individual slot with a tapered T-shaped extrusion below it. This T-shaped extrusion is used to align the sensor slot prior to loading the sensor by funneling it into the T-shaped hole on the gripper adjacent to the sensor slot (Fig. \ref{fig:replacement}(b)). The end of the T-shape extrusion starts with 18 $\%$ of the original surface area and expands. This allows the sensor holder and the gripper to engage with higher tolerance, and as the surface area expends, it aligns the sensor slot perfectly prior to loading the sensor. Without the funneling, the gripper is not capable of the submillimeter precision alignment, often resulting in collision and gripper damage. This funneling approach can be expanded to other application that requires high-precision alignment of the gripper or manipulator for various sensor or tool switching purposes.

\subsection{Replacement Motion Sequence}
For the sensor replacement, the arm initially moves toward the sensor replacement mechanism. The gripper facing sideways moves toward the sensor retrieval box, and the linear actuator is retracted fully to open up the sensor lever. As the sensor slot in the gripper is retracted the sensor wing hits the main metal frame, forcing the sensor out of the slot. The unloaded sensor drops into the box below for retrieval. The gripper moves sideways to intentionally hit the flat surface plate located between the box and the sensor holder to ensure that the sensor is removed from the gripper for the edge cases where the sensor is stuck. Due to the forceful sensor insertions into cornstalks, the sensor often becomes stuck in the sensor slot. Once the slot is emptied out, the gripper reaches to the sensor holder for a new sensor. The sensor is aligned by funneling the T-shaped extrusion. Once the sensor is aligned into the sensor slot, the sensor lever hooks on to the sensor and the arm moves outward to retract the sensor from the sensor holder. The sensor holder contains 5 sensors and therefore can do 5 replacements, but can be easily extended to hold more sensors.

\section{SENSOR MAINTENANCE} \label{sensor_maintenance}

\begin{figure}[t]
    \vspace{1.5mm}
    \centering
    \includegraphics[width=0.95\linewidth]{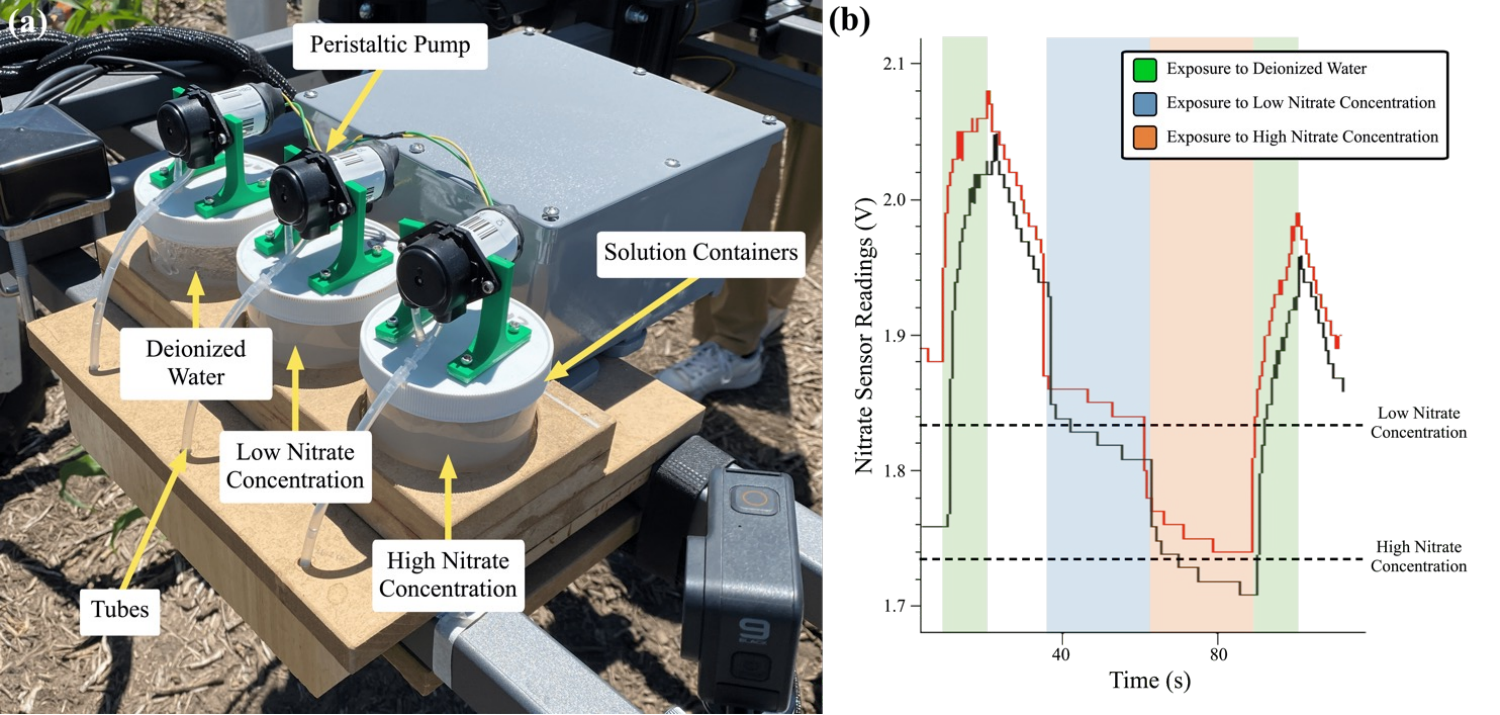}
    \caption{(a) The calibration unit mounted on Amiga robot base. (b) The expected behavior for the nitrate readings during the calibration process shown on two sensors (red and black).}
    \label{fig:calib}
    \vspace{-15pt}
\end{figure}

\begin{figure*}[t]
    \vspace{2mm}
    \centering
    \includegraphics[width=0.9\textwidth]{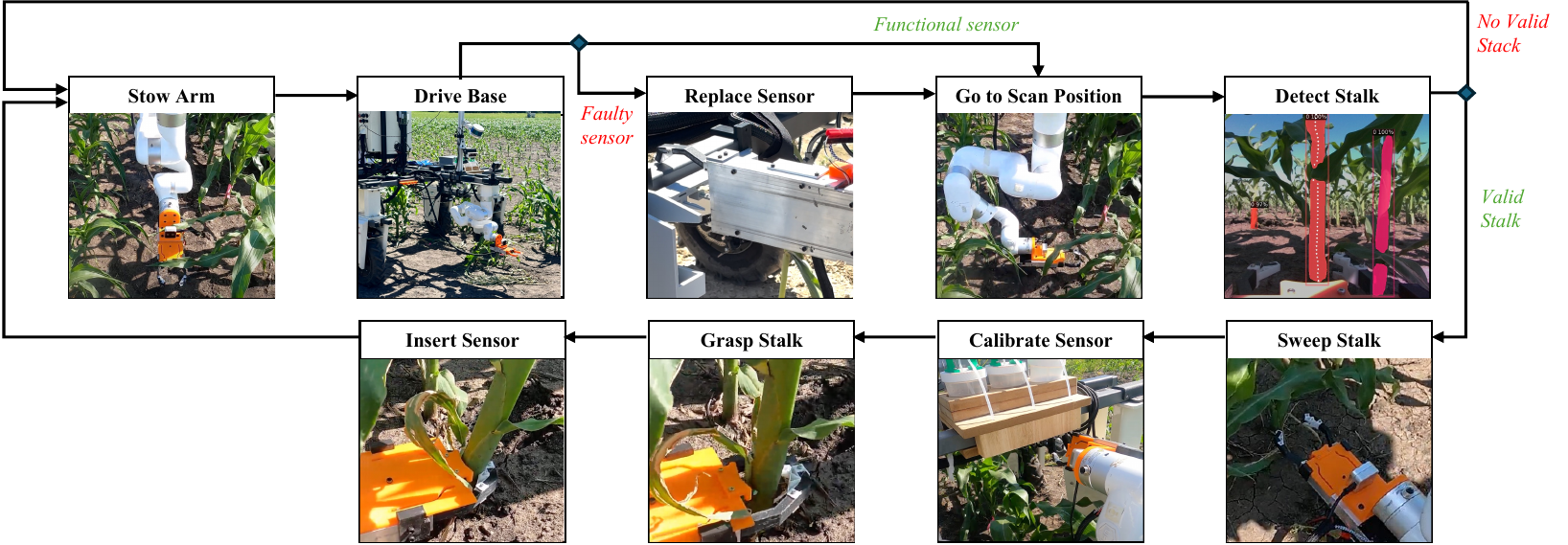}
    \caption{For the sensor insertion task, the robot reaches the sampling location and the sensor is replaced. Then, the stalk is detected for the optimal insertion pose and the sensor is calibrated. The gripper grasps the target stalk and inserts sensor.}
    \label{fig:motion_sequence}
\end{figure*} 

Our nitrate sensor uses an epoxy bioresin that varies in response from sensor to sensor and requires calibration before use \cite{sensor}. Manual monitoring and calibration by humans is a tedious process for interactive sensors \cite{soil2}, but in this work we present an autonomous system for sensor maintenance.

The sensor calibration mechanism, shown in Fig. \ref{fig:calib}(a), consists of three containers holding solutions with different concentrations of nitrate solution: deionized water (0 ppm), 200 ppm and 2000 ppm. Each container has a 5V DC peristaltic pump connected with a tube that provides a dripping flow of solution. For the calibration process, the robotic arm positions the gripper such that the sensor is placed under the extended tubes to ensure the flow of solution directly onto the sensor electrodes.

Nitrate sensor calibration is conducted using two-point calibration given the two solutions of known concentration. The sensor is exposed to the low and high concentration nitrate solutions, and the corresponding voltage readings are recorded (Fig. \ref{fig:calib}(b)). From this calibration, interpolation can be created based on the linear relationship of the nitrate concentration and voltage for this nitrate sensor \cite{sensor}. The sensor is cleansed with deionized water before and after calibration to remove any contaminants.

\section{OPTIMAL INSERTION POSE} \label{pose}
Good sensor maintenance also involves using well aligned insertions to minimize stress on the sensor. As mentioned previously, to ensure accurate readings, the sensor must be inserted at least 8.5mm into the pith region. The likelihood of sensor insertion at a proper depth increases as the cross-sectional view of the cornstalk increases. Due to its ovular nature, the width of a cornstalk varies based on its viewing angle. Hence, we propose using multiple views to inspect the stalk from different angles to determine which angle has the widest surface and is thus the optimal sensor insertion pose.

\subsection{Detection}
 The goal of the visual detection system is to determine the optimal sensor insertion location on the cornstalk in 3D space. This is achieved by detecting the stalks in an RGB-D image from the Intel Realsense D405 stereo camera using a 2D segmentation model, fitting a line to the mask in 3D space, then rejecting invalid stalks. This system serves as an extension of our previous work \cite{pw1}, namely the addition of stalk width estimation to determine the optimal insertion angle with highest success.

The cornstalk width is determined by fitting a 2D line to the RGB mask of the cornstalk, determining the median width across that line in pixels, then using the camera intrinsic and cornstalk depth reading to determine the width in millimeters and ultimately the optimal insertion angle. This implementation was evaluated in the field, and from the 23 successful insertions, 56.52\% of the insertions were within 45 degrees of the ideal insertion angle showing that the majority of insertions were close to the ideal. 


\begin{figure}[t]
    \centering
    \includegraphics[width=0.95\linewidth]{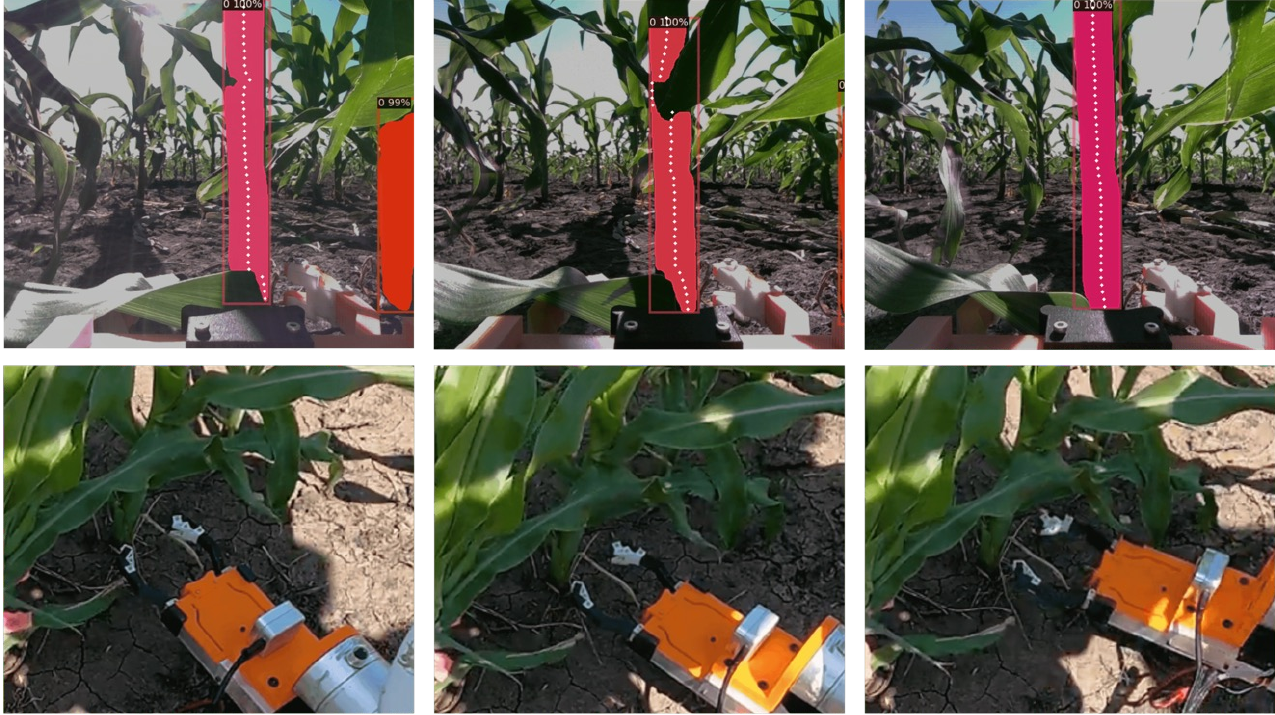}
    \caption{The robotic arm rotates the camera view about the cornstalk to determine which viewing angle has the largest width and thus the highest likelihood of sensor insertion success.}
    \label{fig:sweep}
    \vspace{-15pt}
\end{figure} 

\subsection{Motion}
For the sensor insertion motion (Fig. \ref{fig:motion_sequence}), the xArm robotic arm begins in the stow position which protects the arm and the gripper while navigating the cornfield. Once the sampling location has been reached, the arm moves to the scan position to view the stalks and select the cornstalk with highest probability of success based on distance from the arm, confidence in detection, and width.

To determine the best insertion pose, the arm approaches the selected cornstalk at a height within the pith region. The arm rotates the camera view about the cornstalk in two 15 degree increments, sweeping 30 degrees (Fig. \ref{fig:sweep}). The angle with the largest width is chosen for the final sensor insertion pose. 

During the final steps of the sensor insertion, the robotic arm aligns the center of the gripper with the previously inspected cornstalk. The robotic arm rotates the gripper to the angle of insertion determined by the sweep motion, grasps the fingers around the stalk, and the gripper inserts the sensor. Once the nitrate reading has been recorded from the sensor, the gripper retracts the sensor and opens up the fingers, and the robotic arm returns to the stow position.

\section{EXPERIMENTAL RESULTS}

\subsection{Field Evaluation}

The sensor insertion task was tested during June 2024 at the Iowa Ames Curtis Farm. The cornstalks were at the growth stage of V6-V7, which had the average cornstalk diameter of 21 mm and height of 56 cm. The sensor insertion task was tested on 30 different cornstalks at the field. For the evaluation, the metric consists of 5 criteria involving robot evaluation and manual post-trial evaluation.


\begin{figure}[t]
    \vspace{1.5mm}
    \centering
    \includegraphics[width=0.95\linewidth]{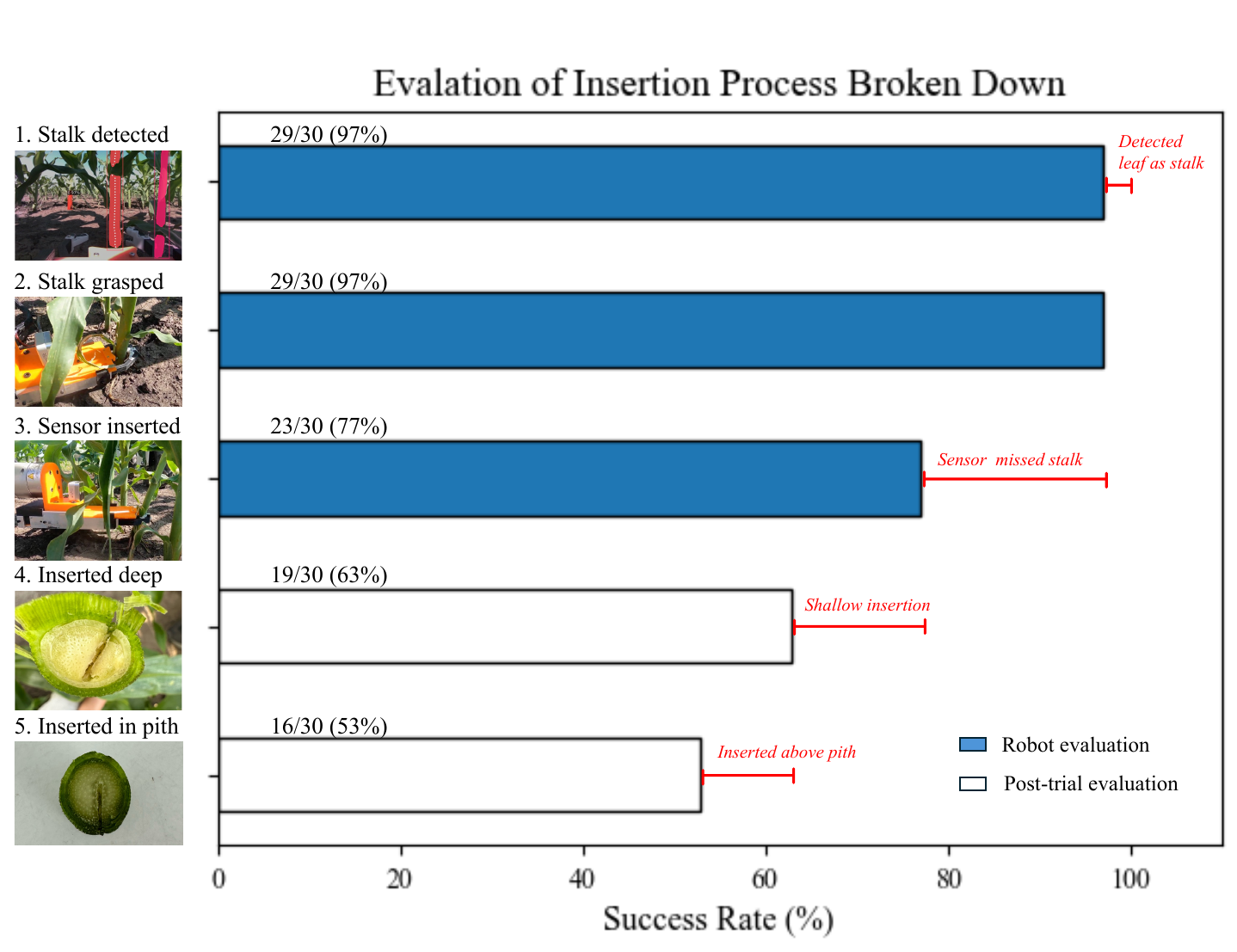}
    \caption{Overall performance of sensor insertion evaluated on 30 field cornstalks.}
    \label{fig:insertion_result}
\end{figure}

\begin{figure}[t]
    \centering
    \includegraphics[width=0.95\linewidth]{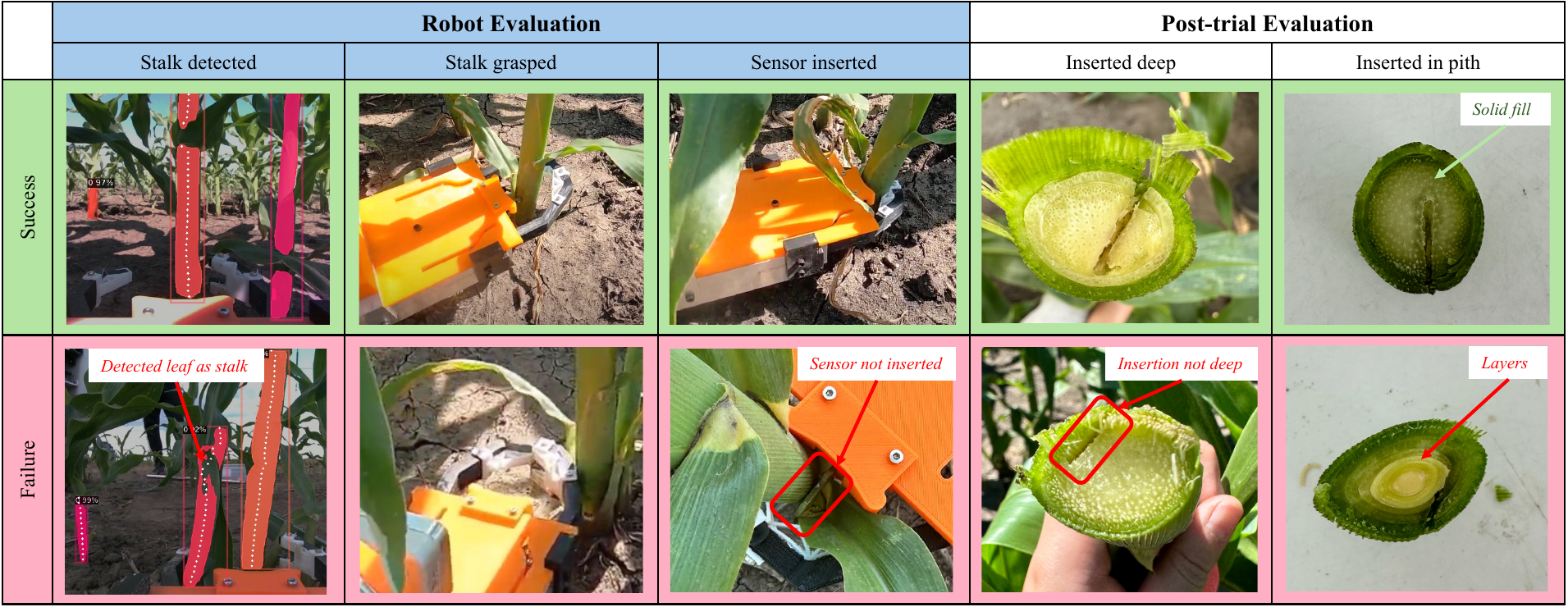}
    \caption{Examples of success and failure cases for each of the criteria for the sensor insertion evaluation.}
    \vspace{-15pt}
    \label{fig:failure}
\end{figure}

As outlined in Section \ref{pose}(B), the insertion motion sequence is mainly detecting a target cornstalk followed by grasping the stalk and inserting the sensor. The success of these three motion steps were evaluated for the robot evaluation. The stalk detection resulted in 97\% success rate, with only a single failure case out of 30 trials where a leaf was detected as a stalk, as shown in Fig. \ref{fig:failure}. Whenever the stalk was detected, the robot was able to grasp the stalk in between the gripper fingers, indicating a successful and reliable arm motion sequence. Out of the 29 cases where the stalk was successfully grasped, the sensor insertion was successful for 23 stalks, resulting in a progressive success rate of 77\%. The cause of the failure cases were the misalignment between the sensor and the center of the stalk. 

After the sensor insertion, it is necessary to determine if the insertion is deeper than 8.5 mm in the pith region. This evaluation was done after the insertion by slicing the cornstalk horizontally at the insertion height. By measuring the depth of the insertion, we were able to verify that there were 19 successful insertions with sufficient depth, resulting in a collective success rate of 63\%. Sensor misalignment with the stalk's center was the primary cause of failures, as shown in Fig. \ref{fig:failure}. Among the 19 successful sensor insertions, 16 sensors were inserted in the pith region, which gives a final overall success rate of 53\% for the sensor insertion. This is an improvement compared to 31\% overall success rate of crop monitoring robot developed by Lee et al. \cite{pw1}.

\subsection{Sensor Replacement Evaluation}

To extensively evaluate the performance of the sensor replacement, the sensor exchange process was tested for 50 iterations. Each trial involved unloading the sensor from the gripper into the retrieval box and loading a new sensor held in the sensor replacement mechanism. The sensor replacement mechanism holds 5 sensors, and each sensor was tested for 10 trials. Out of 50 iterations, the results showed that unloading and loading the sensor is highly reliable with a success rate of 100$\%$.

\subsection{Sensor Calibration Evaluation}
A total of 40 test runs were conducted across 25 sensors to evaluate the sensor calibration unit. Each run involved exposing the sensor to the three solutions for 15 seconds each. Common sensor failure cases are when the two calibration voltages are approximately the same or when the high-concentration nitrate solution is read as a higher voltage when expected to be lower. A successful calibration rate of 62.5\% was determined by whether the sensor exhibited the expected behavior shown in Fig. \ref{fig:calib}(b).

\section{DISCUSSION AND CONCLUSION}

The results of our field evaluation demonstrated a successful robotic system and gripper for sensor insertion and replacement. The two-finger design allowed the gripper to grasp the stalk by moving forward in a straight line with the camera sensor viewing the stalk. When the stalk was successfully detected, the arm manipulation and gripper grasp motion was successful with no failure cases, indicating effective motion sequence and gripper design. The gripper was compact in size using the single-actuator coupled sliding mechanism, and this design allowed the sensor to be inserted very close to the ground to reach the pith region.

The primary cause of insertion failures (10 out of 30 cases) was sensor misalignment that resulted in missed or shallow insertions. Future work should focus on developing a closed loop vision-based robot arm control to center the gripper with high accuracy and precision prior to insertion. 

The sensor replacement process demonstrated the exceptional reliability for unloading and loading sensors, achieving a 100$\%$ success rate. These evaluations suggest that our approach offers a cost-effective and mechanically simple solution for robotic applications demanding high-precision insertions with low-precision manipulators.

The sensor calibration process achieved the desired convergence pattern with a success rate of 63\%. These failures were not evenly distributed and 47\% of these failures occurred in the fourth quarter of testing. This increased failures are due to the expected wear and tear of interactive sensors \cite{sensor} after multiple insertions. This motivates automatic fault detection for future works to systematically detect sensor failures and replace the sensor accordingly. This would provide an increased level of sensor maintenance and system robustness.

\addtolength{\textheight}{0cm}   





\section*{ACKNOWLEDGMENT}

This work was supported in part by NSF Robust Intelligence 1956163 and NSF/USDA-NIFA AIIRA AI Research Institute 2021-67021-35329. The authors would like to thank Vignesh Kumar, Dr. Dong, and Lisa Coffey from Iowa State University for helping with the sensor fabrication process and the field testing at the Curtis Farm. 




\end{document}